\documentclass{article}

\PassOptionsToPackage{numbers, compress}{natbib}

     \usepackage[preprint]{neurips_2020}

\usepackage[utf8]{inputenc} 
\usepackage[T1]{fontenc}    
\usepackage{hyperref}       
\usepackage{url}            
\usepackage{booktabs}       
\usepackage{amsfonts}       
\usepackage{nicefrac}       
\usepackage{microtype}      
\usepackage{floatrow}       

\usepackage{graphicx}

\setcitestyle{square}

\title{Validation and generalization of pixel-wise relevance in convolutional neural networks trained for face classification}

\author{%
  Jñani Crawford\\
  Symbolic Systems Program\\
  Stanford University\\
  Stanford, CA 94305 \\
  \texttt{jnani@stanford.edu} \\
  \And
Eshed Margalit \\
 Neurosciences Institute \\
  Stanford University\\
  Stanford, CA 94305 \\
  \texttt{eshedm@stanford.edu} \\
  \And 
Kalanit Grill-Spector \\
 Department of Psychology \\
  Stanford University\\
  Stanford, CA 94305 \\
  \texttt{kalanit@stanford.edu} \\
  \And
Sonia Poltoratski \\
Department of Psychology\\
  Stanford University\\
  Stanford, CA 94305 \\
  \texttt{sonia09@stanford.edu} \\
}

\begin{document}

\maketitle

\begin{abstract}
 The increased use of convolutional neural networks for face recognition in science, governance, and broader society has created an acute need for methods that can show how these 'black box' decisions are made. To be interpretable and useful to humans, such a method should convey a model's learned classification strategy in a way that is robust to random initializations or spurious correlations in input data. To this end, we applied the decompositional pixel-wise attribution method of layer-wise relevance propagation (LRP) to resolve the decisions of several classes of VGG-16 models trained for face recognition. We then quantified how these relevance measures vary with and generalize across key model parameters, such as the pretraining dataset (ImageNet or VGGFace), the finetuning task (gender or identity classification), and random initializations of model weights. Using relevance-based image masking, we find that relevance maps for face classification prove generally stable across random initializations, and can generalize across finetuning tasks. However, there is markedly less generalization across pretraining datasets, indicating that ImageNet- and VGGFace-trained models sample face information differently even as they achieve comparably high classification performance. Fine-grained analyses of relevance maps across models revealed asymmetries in generalization that point to specific benefits of choice parameters, and suggest that it may be possible to find an underlying set of important face image pixels that drive decisions across convolutional neural networks and tasks. Finally, we evaluated model decision weighting against human measures of similarity, providing a novel framework for interpreting face recognition decisions across human and machine.
  
\end{abstract}

\section{Introduction}
Faces are of tremendous social and developmental importance for humans, and their recognition is carried out by a network of specialized and selective brain areas beyond the early visual hierarchy \cite{grill2017functional}. Effective face recognition requires the discrimination of individual exemplars of an object class that is largely consistent in the presence and relative position of its features (e.g. eyes, mouth, and nose). This makes face recognition an especially challenging problem in neural computation that builds upon general object recognition. Convolutional neural networks (CNNs) imitate the neural mechanics of the human visual system in their hierarchical structure, linear-nonlinear operations, spatial pooling, and emergent complex representational structure. CNNs trained to categorize natural images yield early-layer filters that resemble the orientation-selective spatial receptive fields of early visual cortex \cite{hubel1962receptive,lau2002computational}. Under the same goals of image classification, CNN features at intermediate layers map exceptionally well to neurons in mid- and high-level visual cortex \cite{yamins2016using}, even without any explicit constraints to fit neural data. These successes have made hierarchical CNNs a very popular model of the human visual system, especially as their performance on recognition now often surpasses human ability \cite{phillips2014comparison}. However, relatively little is known about the response characteristics of face-specific, rather than object-general, CNNs and how they may correspond to human face recognition behavior. At the same time, neural-network-enabled face recognition has gained use and prominence in broader society, making it critical for us to understand how networks make decisions and whether they do so in a manner that is consistent with the humans who use them. Interpretability of face recognition models' decisions allows their users to ensure ethical applications, and to avoid technological susceptibilities such as adversarial attacks.

\subsection{Interpretability of CNN decisions}

Several complementary methods have been proposed to explain CNN decisions. Perturbation-based methods measure how changes in input pixels alter the output decision and are particularly useful for highlighting common vulnerabilities, such as weakness to adversarial examples \cite{fong2019explanations}. Direct visualization of convolutional kernels is possible for the input layer of a CNN (e.g. Figure \ref{DatasetExamples}); at later layers, methods such as feature or optimal stimuli visualization \cite{olah2017feature} attempt to approximate unit selectivity by generating example images that maximally activate a given set of units. To provide interpretable visualizations of CNN decisions through the entire hierarchy of layers, attribution methods (e.g. salience or relevance mapping) seek to identify pixels in a given input which are most responsible for the network's decision \cite{adebayo2018sanity,zeiler2014visualizing}. In this work we utilize layer-wise relevance propagation (LRP), which uses the weighted graph structure of the CNN to backpropagate relevance from the logit (pre-softmax activation in final layer) of a particular class in the network's prediction layer back to the input pixels. This relevance is recursively distributed according to each input unit's calculated contribution, which is a linear combination of that unit's input weight and pre-activation \cite{bach2015pixel}. For a given model and input image, LRP produces a relevance map wherein large positive values indicate pixels that contributed towards the network's predicted likelihood of that class, and negative values against \cite{montavon2019layer}. As a decomposition of the activity of the logit of a chosen class, LRP is a straightforward and lightweight method, as it requires no additional training or optimization.

\subsection{Possible driving factors for CNN classification strategies}

The current work evaluates the contribution of three factors that may influence model decision processes: (i) the content of the training dataset \cite{torralba2011unbiased}, (ii) the nature of the loss function or task \cite{ozbulak2016transferable}, and (iii) the effects of random initializations of the re-learned readout layer added prior to finetuning.  We first verify that the relevant pixels of face images as selected by LRP are in fact important to CNN classification accuracy by removing relevant pixels and evaluating task performance. Then, we use the same method to quantify the generalizability of LRP relevance across several instances of VGG-16 CNNs that vary in their pretraining dataset (ImageNet or VGGFace), finetuning task (gender or identity classification), and five random finetuning initializations. Evaluating the generalizability of LRP-derived relevance maps provides critical knowledge about the decisions made by CNNs; specifically, lack of generalizability would imply that the decision process of a face-recognizing system is critically sensitive to design decisions made by those deploying these networks.

\section{Methods}

We evaluated 20 different VGG-16 CNNs  implemented in Keras \cite{chollet2015keras}. The models were pretrained on one of two datasets (10 each): ImageNet-1000 classification or the VGGFace dataset. These models were then finetuned on the 1,515 greyscale face images from 101 identities in the Vision and Perception Neurscience Lab face dataset (VPNL-faces). Two finetuning tasks were used: models were trained to classify either gender or identity. Five models were trained for each combination of pretraining and finetuning tasks (four total; Figure \ref{ModelConditions}a). Subsequently, LRP was used to derive pixel-wise maps of relevance for the 101 front-facing oriented face images of each person in the VPNL-faces dataset. To validate the relevance measure and evaluate generalization across model parameters, we occluded the top Nth-percentile of the pixels by absolute relevance value and evaluated the models on these masked images. This yielded the percent accuracy of the model on the finetuning task as a function of the percentage of masked pixels. 

\subsection{Pretrained models and datasets}

\begin{figure}[htp]
  \centering
  \includegraphics[width=13.5cm]{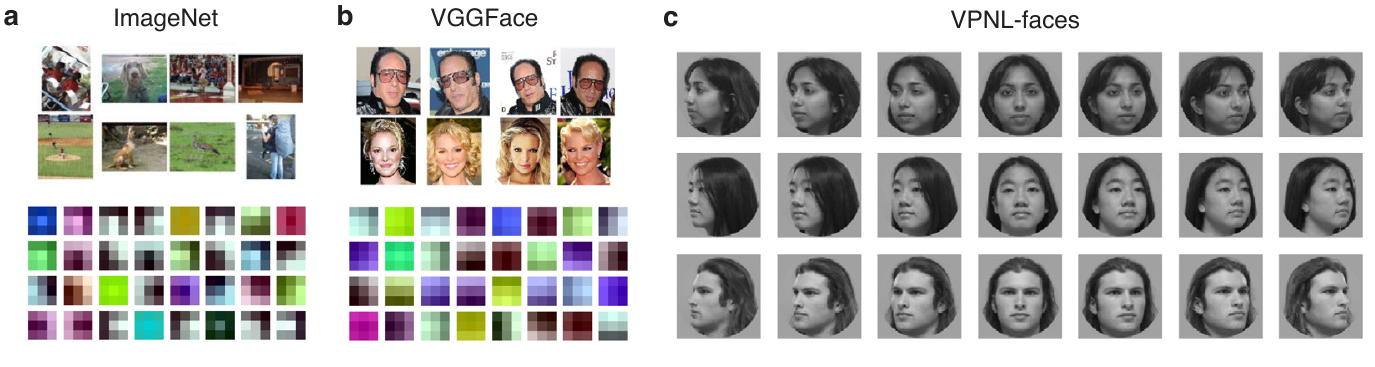}
  \caption{Details of the datasets and resulting kernels from the first convolutional layer for \textbf{a.} VGGFace-trained networks and  \textbf{b.} ImageNet-trained networks. \textbf{c.} Sample images from VPNL-faces dataset used for finetuning, which was comprised of 1,515 grayscale cropped photos of 101 members of a university community (44 labeled male and 57 female) with viewpoints at 15 degree increments from -105 to +105 along the vertical axis.}
  \label{DatasetExamples}
\end{figure}

The VGG-16 architecture was chosen because it can achieve near-perfect accuracy on the finetuning dataset and are compatible with the LRP methods used, whereas some other popular model architectures (e.g. ResNet-50) have pooling layers that lead to artifacts in the computed relevance maps. The ImageNet-trained models come from the Keras models package, and are trained on the standard ImageNet-1000 classification challenge as described in \citep{simonyan2014very} and shown in Figure \ref{DatasetExamples}a. The VGGFace-trained models were converted from the original Caffe models from the Oxford Visual Geometry Group \cite{parkhi2015deep} and made available as Keras models at \cite{rcmallikerasvggface}. These were trained on the VGGFace dataset described in Figure \ref{DatasetExamples}b. To match the architecture of the VGGFace-trained network to the ImageNet-trained network, we removed the final two fully-connected layers. This removal did not prevent the model from reaching near-perfect accuracy on either of the finetuning tasks. 

The ImageNet dataset has the advantage of a diverse image and label set, while the VGGFace task requires more fine-grained discrimination between many exemplars of human faces. This may better enable the construction of a high level 'face space' that can encode the various important dimensions for robust face recognition.\cite{o2018face}. We can see from visualizing the kernels from the first convolutional layer of both the ImageNet and VGGFace pretrained weights in Figure \ref{DatasetExamples} that the ImageNet kernels and the VGGFace kernels, that both have oriented, color-selective 'edge' filters at their first layer. However, in accord with its more specialized training set, the VGGFace filters appear to be lower spatial frequency and sample a more limited color range. It is unresolved if the types of visual input modeled by either of these datasets are necessary or sufficient for face recognition behavior in humans,  but either may appropriately model at least a portion of visual input during human development.

\subsection{Finetuning}

Each of the 20 models was finetuned on one of two tasks using the VPNL-faces dataset \cite{srivastava2017training}, described in Figure \ref{DatasetExamples}c. The images were preprocessed similarly to training images, by subtracting the mean value of each ImageNet color channel from each pixel. Finetuning involved freezing the weights of all convolutional layers, resizing the final fully connected layer to the 101 identities in the dataset or the 2 labeled gender classes, and then training on either a 101-way identity classification or a 2-way gender classification. During finetuning, 80\% of the images were used for training and 20\% for validation. All models, regardless of pretraining dataset or finetuning task were trained for 25 epochs and achieved near-perfect performance on the VPNL-faces dataset, with better than 98\% accuracy in all cases. 

\subsection{LRP maps}

Layerwise relevance propagation was done via composite LRP \cite{montavon2019layer} as implemented in Alber and Lapushkin’s python library \textit{iNNvestigate} \cite{alber2019innvestigate}. Each of the 20 models yielded a relevance map for each of the 101 front-facing oriented VPNL-faces; the resulting maps were normalized to the range [-1,1]. To measure similarity between maps generated from different models for the same face image, we computed the Pearson's coefficient between vectorized relevance maps.

\subsection{Validating relevance maps with masking}

\begin{figure}[htp]
  \centering
  \includegraphics[width=13.5cm]{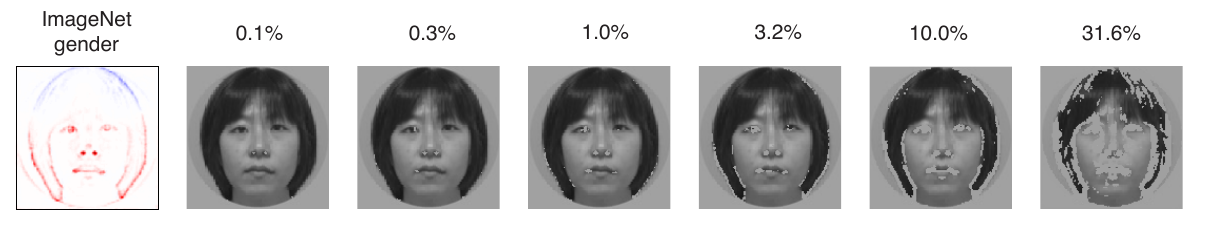}
  \caption{Example of masking $N$th percentile of pixels an image from the VPNL-faces dataset randomly and using one of the models' relevance maps, for increasing values of $N$.}
  \label{MaskingExample}
\end{figure}

To assess the importance of high-relevance pixels toward a model's classification decisions, we performed iterative masking from the relevance maps (Figure \ref{MaskingExample}). The top Nth-percentile of pixels in the 101 front-facing VPNL-faces dataset images were removed by corresponding absolute value of relevance; the model was then re-evaluated on the masked images. If the LRP method is valid, we expect that increasing the proportion of masked pixels should reduce performance to chance on the finetuning task.

\begin{figure}
\floatbox[{\capbeside\thisfloatsetup{capbesideposition={right,top},capbesidewidth=4cm}}]{figure}[\FBwidth]
{\caption{Representative pairings from each of the five types of pairwise model comparisons (bins) made to test the effects of pretraining dataset, finetuning task, instantiation, or multiple of these in combination.}\label{ModelConditions}}
{\includegraphics[width=9cm]{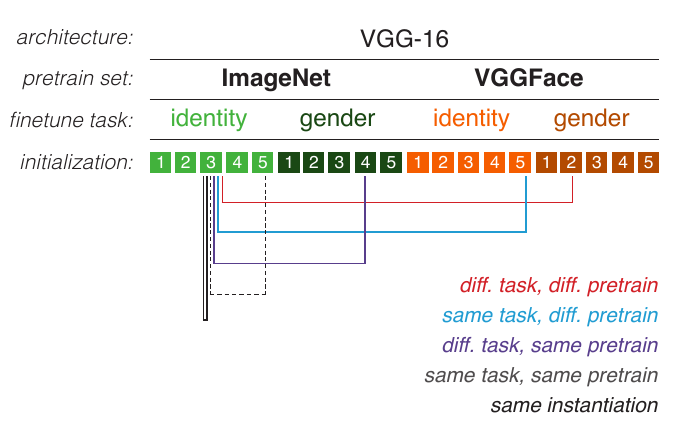}}
\end{figure}

The generalizability of the LRP method is assessed by applying masks made from the relevance maps of one 'source' model to some other 'destination' model. Each pair of source or destination models may differ by pretraining dataset, finetuning task, and/or random instantiation of finetuning weights. This yields 5 types of comparisons, or 'bins' (Figure \ref{ModelConditions}b). For each of the 20 possible destination models, a source model was chosen with replacement 100 times per bin. To assess the stability of relevance maps across random associations, we compared performance curves between 'same instantiation' (solid lines) and 'same task + same training' (dotted lines) in Figure \ref{SamplingCurves1}. The effects of pretraining and finetuning task were similarly assessed (Figure \ref{SamplingCurves2}). 

\section{Results}

\begin{figure}
  \centering
  \includegraphics[width=13.5cm]{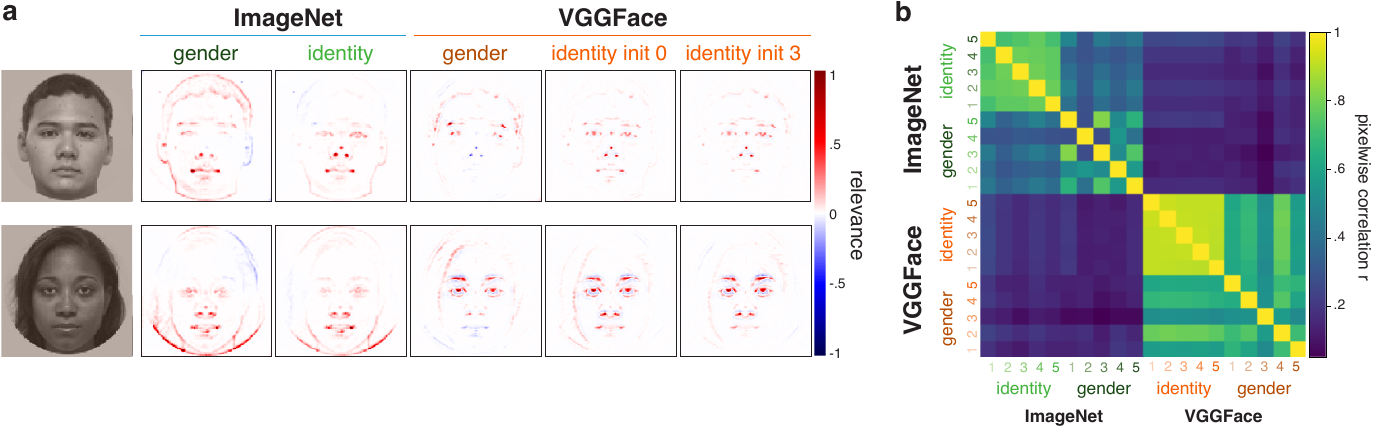}
  \caption{\textbf{(a)} Sample images and relevance maps for 2 identities, from 5 different models. Positive values are pixels which contribute towards the predicted class. Negative values are pixels which contribute against the predicted class. Zero valued pixels have either no or contradictory contributions. \textbf{(b)} Similarity matrix for the relevance maps from all 20 models. For each pair of models, the similarity measure is the Pearson's correlation coefficient between the pixels of vectorized relevance maps for each of 101 front-view face images.}
  \label{RelevanceExamples}   \label{SimilarityMatrix}
\end{figure}

Examples of relevance maps generated for different models are shown in Figure \ref{RelevanceExamples}. The relevance maps of all models seem to highlight meaningful face features. Qualitatively, the most discernible difference is the greater reliance on internal features of the face (e.g. eyes, nose) in VGGFace-trained models than in ImageNet-trained models, as well as more fine-grained positive/negative relevance combinations in these regions. ImageNet-trained models, instead, assign more relevance to external features like the hair, particularly in discriminating gender. As evidenced in the two right-most panels, relevance maps are largely unchanged across random instantiations of the same training/task model. 
To evaluate whether similar relevance maps were computed from models across pretraining, finetuning task, and initializations, we computed Pearson's correlations for maps generated on the same face image. The average pixelwise similarity of maps of the 101 face images from the 20 models are summarized in Figure \ref{SimilarityMatrix}. We see that models with the same pretraining yield the most similar relevance maps, and that relevance within VGGFace-trained models is more similar across tasks than within ImageNet-trained models. ImageNet-trained models finetuned on gender classification show the largest variance in relevance maps generated from models with different random fine-tuning weight initialization.

\subsection{Effects of relevance-based occlusion on model accuracy and stability over random initialization.}

\begin{figure}
  \includegraphics[width=12cm]{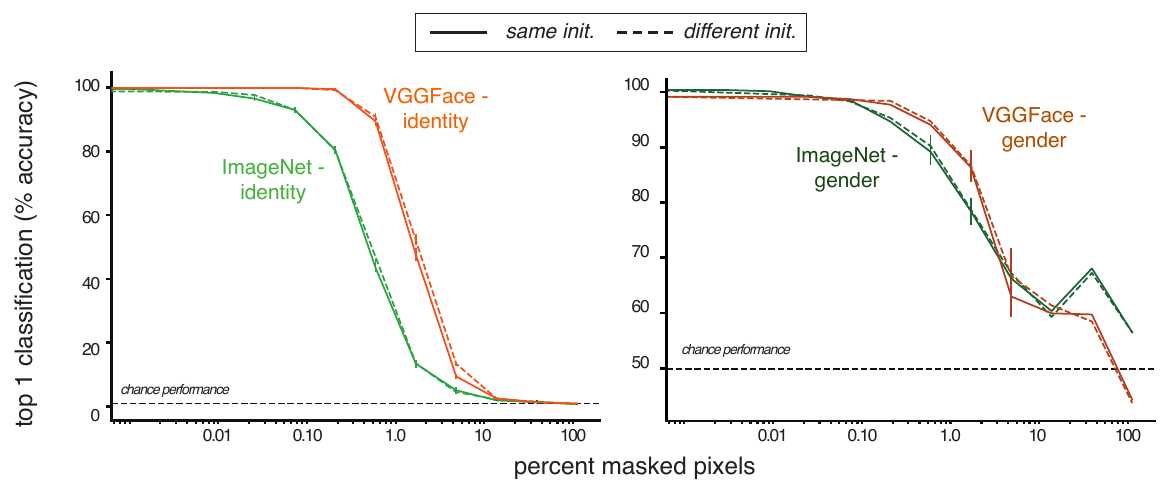}
  \caption{Model performance on images masked according to relevance maps generated from the destination model (same init.) or a different instantiation of model with the same pretraining and finetuning task. \textbf{a.} Models finetuned on VPNL-face identity classification. \textbf{b.} Models finetuned on VPNL-face gender identity classification.}
  \label{SamplingCurves1}
\end{figure}

First, we evaluated how relevance-based masking hurts model performance as the number of masked pixels increases, a measure of the validity of the relevance maps toward the classification decisions. The masked model performance for the four pretraining/task combinations of models is shown in Figure \ref{SamplingCurves1}. The solid curves show the results of applying masks generated from a given model to the inputs of that exact model ('same init.'). In all cases, model accuracy approaches chance as more pixels are occluded, but performance appears more resilient to occlusion in VGGFace-trained models (orange curves). Within each training regime, accuracy drops more sharply to chance in the identity classification task than in gender classification. At one masking level (31.6\% of pixels) there is a small increase in performance in ImageNet-trained gender classification models (green curves). This corresponds to asymmetries in positive/negative relevance value distributions in these maps: because pixels are iteratively removed according to their magnitude (but not sorted by sign), positive values may be removed before negative values. Thus, pixels remaining at higher thresholds are more likely to correspond to negative relevance, and in a binary classification task, negative relevance is expected to skew predictions to the incorrect label.

To quantify the stability of LRP-generated relevance maps across different random initializations of finetuning weights, we evaluated performance for a given 'destination' model using masks from 'source' models with the same pretraining and finetuning task, but different initial readout weights. We quantified the resilience of each destination model to relevance-based masking by computing the area-under-the-curve (AUC) for different source models. Computing AUC has three primary benefits: it i) allows us to assess accuracy falloff with increasing percentages of masked pixels without fitting curves to the data, ii) ensures stability of results given reasonably-fine sampling of percentages, and iii) has a clear interpretation: higher AUC indicates more resilience to masking (e.g. smaller drop in classification performance). For each finetuning task, a two-way repeated-measures ANOVA was run to test the effect of mask source ('same-init' vs. 'different-init.') and pretraining dataset (ImageNet vs. VGGFace) on AUC. 
For the identity task (Figure \ref{SamplingCurves1}, left panel), we found a main effect of pretraining dataset (F(1, 16) = 49.6, $p$ = 2.8e-6), with AUC significantly higher for VGGFace-trained networks than for ImageNet-trained networks. We found no main effect of mask source (F(1, 16) = 1.88, $p$ = 0.19), indicating that AUC was comparable between same-init and different-init. sources. 
We found the same pattern of results in the gender task (Figure \ref{SamplingCurves1}, right panel): a significant main effect of pretraining dataset (F(1, 16) = 34.5, $p$ = 2.4e-5), and no effect of same-init vs. different-init. (F(1, 16) = 0.09, $p$ = 0.77). Together, these results demonstrate that for both finetuning tasks, VGGFace-trained networks were more resilient to masking than ImageNet networks, and that the random initialization of finetuning weights did not have a significant effect on resilience as quantified by AUC. Given differences in AUC based on the pretraining dataset, we analyzed the contribution of pretraining and finetuning separately in further analyses.

To evaluate the effect of both finetuning task and pretraining dataset on a given model's AUC, we compared the effects of masking from four source models: same-init (orange curves in Figure \ref{SamplingCurves2}), different pretraining and different finetuning task (red curves), different pretraining but the same finetuning task (cyan curves), and the same pretraining but different finetuning task (violet curves). In all four cases, a repeated-measures ANOVA revealed a significant main effect of mask source on the AUC of the destination model (all F(3, 16)s > 3.3, all $p$s < 0.05), indicating that the effect of relevance-based masking on model performance does not generalize across all combinations finetuning and pretraining datasets. In the following sections, we use multiple-comparisons-corrected tests to unpack the differences in AUC of curves obtained with different pairs of mask sources.

\subsection{Strong relevance generalization across finetuning task}
\begin{figure}
  \includegraphics[width=12cm]{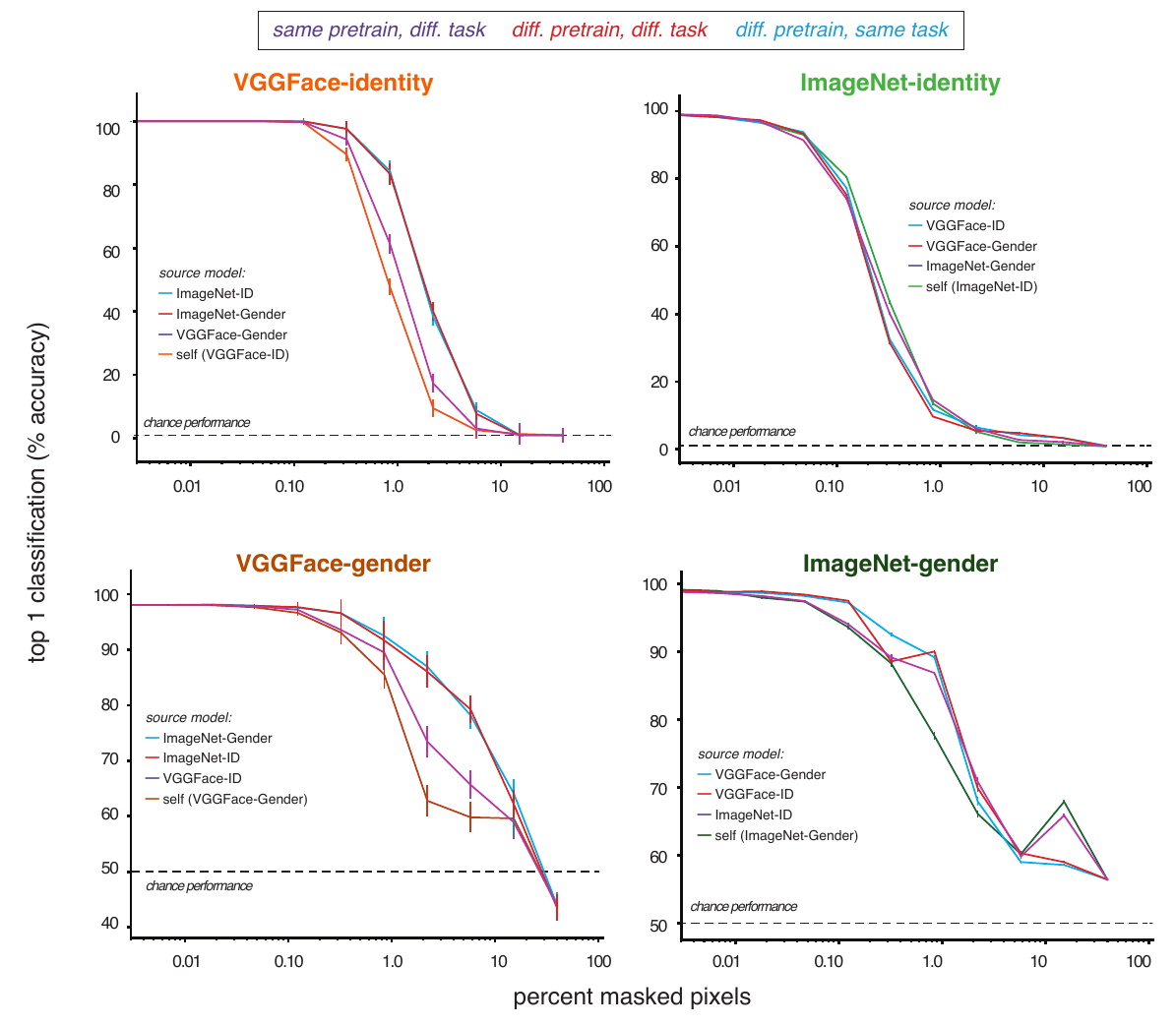}
  \caption{Model performance for classifying relevance-based masked images, split by destination model pretraining dataset and finetuning task.}
  \label{SamplingCurves2}
\end{figure}

We next evaluated the degree to which the finetuning process, and thus, the task that the model is ultimately performing, affects the measured relevance of pixels in the input images. To do so, we asked whether the effect of masking significantly decreased when creating masks from models pretrained on the same task as the destination model but finetuned on a different task, as compared to the same-init model performance. 
Across the four destination model types in Figure \ref{SamplingCurves2}, we observe similarity between performance curves for masks obtained from the same-init (orange curves) and masks obtained from models trained on different finetuning tasks (violet curves). Pairwise tests confirm that in all four cases, the AUC for these two curves was indistinguishable, all $p$s > 0.06. This result is consistent with the fact that the finetuning process modifies weights in a single layer of the network, thus affecting a small proportion of the model's parameters. Next, we asked whether the effect of relevance-based masking differed between models with different pretraining, or whether all models sharing a common architecture are comparable with respect to the input features they rely on to make their decisions.

\subsection{Weaker generalization across pretraining datasets}

If relevance-based masks generalized across pretraining datasets with respect to their effect on destination model accuracy, it would suggest that the feature basis learned during pretraining is similar enough that masks generated from either ImageNet- or VGGFace-trained models would be functionally indistinguishable. Alternatively, failure to generalize would suggest that fundamentally different features are being learned during pretraining, and thus, that the models are weighting input features differently to accomplish the downstream task.

We found that generalization across pretraining datasets is substantially weaker than the robust generalization across random initialization and finetuning tasks we report above. For three of the model destination types (ImageNet-trained-identity-finetuned, VGGFace-trained-identity-finetuned, and VGGFace-trained-gender-finetuned), AUC was significantly higher when using masks obtained from models with different pretraining datasets (cyan and red curves) than when using masks obtained from the same-init model (orange curve; all $p$s < 0.03). While both different-pretraining-same-finetuning and different-pretraining-different-finetuning deviated from the same-init model, their effect on AUC was statistically indistinguishable from each other for all four destination model types (all $p$s > 0.92), consistent with our findings that finetuning task does not affect mask efficacy.
For the Imagenet-trained-gender-finetuned model, AUC did not differ across pretraining datasets (both $p$s > 0.08). We speculate that the divergence of this destination model type from the other three may be due to the non-monotonic shape of the accuracy curves that is not observed in the other cases. Taken together, these results demonstrate generalizability across finetuning tasks and random initializations, but not across pretraining datasets.

\subsection{Relevance interpretability and comparisons to human observers}

\begin{figure}
  \includegraphics[width=13.5cm]{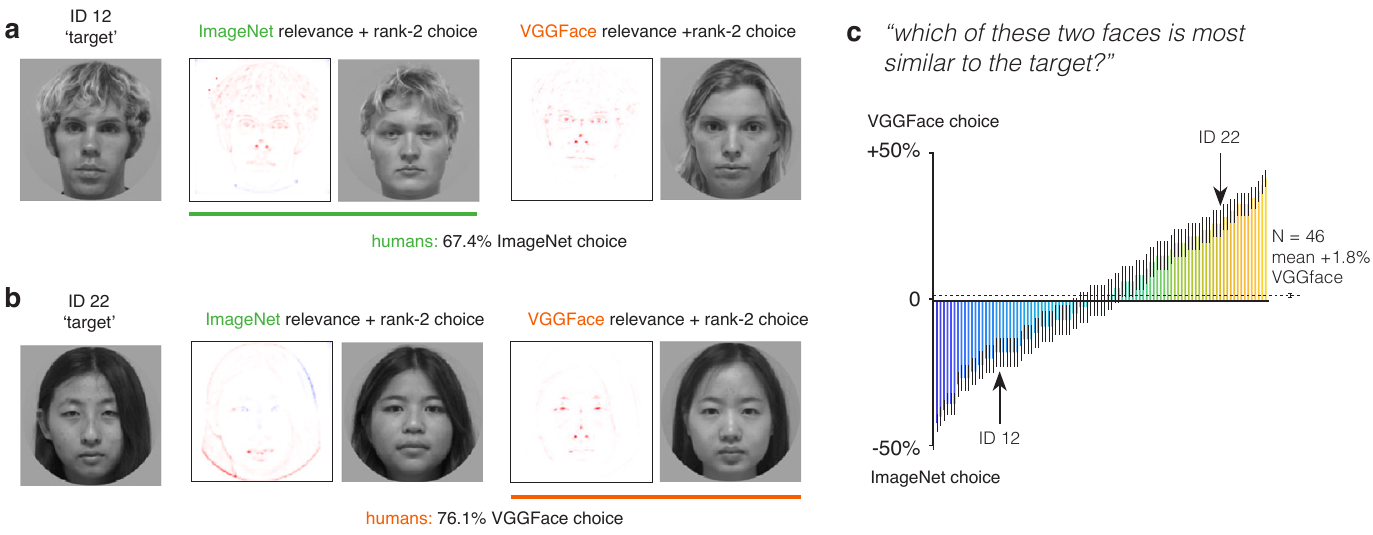}
  \caption{Model similarity metrics via rank-2 choice visualization and comparison to humans \textbf{a.} A target identity (rank-1 choice) alongside rank-2 choices for ImageNet- and VGGFace-trained models. In this instance, humans judge the ImageNet-chosen face to be most similar to the target. \textbf{b.} An exemplar in which humans consistently choose the VGGFace-trained model rank-2 face as most similar to the target. \textbf{c.} Ranked choice results for human similarity judgements.}
  \label{HumanData}
\end{figure}

We next sought to develop a lightweight testing framework to determine the correspondence between CNN information sampling toward face classification and human face recognition behavior. While it may be possible to directly measure human recognition of relevance-based masked images, doing so requires a calibrated experimental environment and time-intensive behavioral measurements. Instead, we first identified a complementary source of information about CNN classification strategy: the rank ordering of output layer activations in response to each face identity. For each face identity ‘target’ that had the highest output layer activation at the decision layer (rank-1 logit), we considered the identity of the face image with the next-highest output activation (rank-2 logit) as an approximation of which identity each model considered 'most similar' to the target face. Doing so in conjunction with relevance mapping provided an additional view on the strategies used by our face-classifying CNNs: in Figure \ref{HumanData}a, for example, the ImageNet-trained model assigns more relevance to pixels over the hair of the target image, while the VGGFace-trained model weights internal assigns maximal relevance to regions of the internal features, including the nose. This preference is easily apparent from the rank-2 choice images of each network, which resemble the target face preferentially in the relevance-weighted regions.

Given this measure of 'most similar' faces in the VGGFace- and ImageNet-trained CNNs for face identification, we asked whether either human observers judged similarity between the faces in a manner that was more consistent either class of models. To do so, we identified 95 face identities from the full 101-face VPNL dataset on which the ImageNet- and VGGFace-trained models produced differing rank-2 choices. We asked human observers (N = 46) on Amazon Mechanical Turk to choose which identity was most similar to the target. Overall, there was a modest but significant preference for the VGGFace rank-2 choice (51.8\% +/- 0.77, t(45) = 2.345, $p$ = 0.023). While across the faces, human preference was nearly equivalent between VGGFace and ImageNet rank-2 choices, we see strikingly consistent human choice preference for many individual identities (Figure \ref{HumanData}c). From this, it appears that human observers use both ImageNet- and VGGface-like information sampling to judge similarity, relying at times on external features like hair shape (e.g. Figure \ref{HumanData}a), and at other times on fine-grained discrimination of face features (e.g. Figure \ref{HumanData}b). Indeed, this is consistent with neural substrates of face recognition in humans: while humans have cortical regions specialized for high-level visual recognition, these are situated at the end stage of a general hierarchy for visual object recognition. The current result underscores the importance of considering both general ‘low-level’ and face-specific mechanisms in modeling how humans distinguish face identity in naturalistic settings as represented in the VPNL-faces dataset.

\subsection{Summary}

Occluding the relevant pixels identified by LRP generally had a negative effect on a model's accuracy, validating the importance of these pixels for performance. The rate of accuracy degradation with increasing masking percentage is not the same between the different types of models: VGGFace-trained models were generally more tolerant to the occlusion than ImageNet-trained models.  Relevance maps produced by various types of VGG-16 models prove stable across random initializations, suggesting that these models are not overly reliant on spurious correlations in the data, and can generalize well across finetuning tasks. However, it appears that pretraining dataset has a substantial effect on the learned features of networks, and results in a differential weighting of input features toward task performance. Comparing VGGFace- and ImageNet-trained models to human behavior provided evidence that both models make 'human-like' similarity judgments on a subset of face images, and that humans use both sampling strategies in naturalistic recognition of faces.

\section{Broader Impact}

Our results show that relevance maps generated via LRP can localize the most relevant pixels of a face image for CNN classification with some degree of invariance to factors like training data or task. This suggests that it may be possible to derive these relevance maps once for a given network to find patterns of relevant features that may generalize to other face-classification networks. A measure of that generalization strength, as computed here, would prove useful to those seeking to selectively modify image features in order to boost or limit the effectiveness of such networks. For example, someone seeking to anonymize a large database of face images with minimal image distortion may be able to use such relevance maps to apply blurs or masking that efficiently hinder CNN-based identification. One may also be able to use pixel-wise relevance for the opposite effect, perhaps in order to compress a dataset of face images in such a way as to preserve the most relevant pixels while lowering quality in irrelevant image regions.

LRP also provides a method for investigating the similarity of decision-making processes in humans and CNNs performing face-classification. In the case of the VGGFace- vs ImageNet-trained models in this paper, the more robust relevance maps generated by VGGFace may be indicative of a more robust facial encoding, and a measurement of this robustness and tolerance to occlusion is a very useful metric to have in addition to pure performance in terms of accuracy. If they can lead to these kinds of robust relevance maps, it is possible that exposure to a higher number of relevant exemplars during pretraining, while not necessary for accuracy on the finetuning dataset in this case, is the kind of measure necessary to create more secure, trustworthy, and more human-like face classifiers.

\begin{ack}
Funding in direct support of this work: NSF Grant (BSC-1756035) to K. G.-S. and NSF Graduate Research Fellowship to E.M.
\end{ack}

\medskip

\small

\bibliographystyle{acm}
\bibliography{main}

\begin{thebibliography}{10}

\bibitem{adebayo2018sanity}
{\sc Adebayo, J., Gilmer, J., Muelly, M., Goodfellow, I., Hardt, M., and Kim,
  B.}
\newblock Sanity checks for saliency maps.
\newblock In {\em Advances in Neural Information Processing Systems\/} (2018),
  pp.~9505--9515.

\bibitem{alber2019innvestigate}
{\sc Alber, M., Lapuschkin, S., Seegerer, P., H{\"a}gele, M., Sch{\"u}tt,
  K.~T., Montavon, G., Samek, W., M{\"u}ller, K.-R., D{\"a}hne, S., and
  Kindermans, P.-J.}
\newblock innvestigate neural networks.
\newblock {\em Journal of Machine Learning Research 20}, 93 (2019), 1--8.

\bibitem{bach2015pixel}
{\sc Bach, S., Binder, A., Montavon, G., Klauschen, F., M{\"u}ller, K.-R., and
  Samek, W.}
\newblock On pixel-wise explanations for non-linear classifier decisions by
  layer-wise relevance propagation.
\newblock {\em PloS one 10}, 7 (2015).

\bibitem{chollet2015keras}
{\sc Chollet, F., et~al.}
\newblock Keras.
\newblock \url{https://keras.io}, 2015.

\bibitem{fong2019explanations}
{\sc Fong, R., and Vedaldi, A.}
\newblock Explanations for attributing deep neural network predictions.
\newblock In {\em Explainable AI: Interpreting, Explaining and Visualizing Deep
  Learning}. Springer, 2019, pp.~149--167.

\bibitem{grill2017functional}
{\sc Grill-Spector, K., Weiner, K.~S., Kay, K., and Gomez, J.}
\newblock The functional neuroanatomy of human face perception.
\newblock {\em Annual Reviews of Vision Science 3\/} (2017), 167--196.

\bibitem{hubel1962receptive}
{\sc Hubel, D.~H., and Wiesel, T.~N.}
\newblock Receptive fields, binocular interaction and functional architecture
  in the cat's visual cortex.
\newblock {\em The Journal of physiology 160}, 1 (1962), 106--154.

\bibitem{lau2002computational}
{\sc Lau, B., Stanley, G.~B., and Dan, Y.}
\newblock Computational subunits of visual cortical neurons revealed by
  artificial neural networks.
\newblock {\em Proceedings of the National Academy of Sciences 99}, 13 (2002),
  8974--8979.

\bibitem{rcmallikerasvggface}
{\sc Malli, R.~C.}
\newblock keras-vggface: Vggface implementation with keras framework.
\newblock \url{https://github.com/rcmalli/keras-vggface}, 2019.

\bibitem{montavon2019layer}
{\sc Montavon, G., Binder, A., Lapuschkin, S., Samek, W., and M{\"u}ller,
  K.-R.}
\newblock Layer-wise relevance propagation: an overview.
\newblock In {\em Explainable AI: Interpreting, Explaining and Visualizing Deep
  Learning}. Springer, 2019, pp.~193--209.

\bibitem{olah2017feature}
{\sc Olah, C., Mordvintsev, A., and Schubert, L.}
\newblock Feature visualization.
\newblock {\em Distill 2}, 11 (2017), e7.

\bibitem{ozbulak2016transferable}
{\sc Ozbulak, G., Aytar, Y., and Ekenel, H.~K.}
\newblock How transferable are cnn-based features for age and gender
  classification?
\newblock In {\em 2016 International Conference of the Biometrics Special
  Interest Group (BIOSIG)\/} (2016), IEEE, pp.~1--6.

\bibitem{o2018face}
{\sc O’Toole, A.~J., Castillo, C.~D., Parde, C.~J., Hill, M.~Q., and
  Chellappa, R.}
\newblock Face space representations in deep convolutional neural networks.
\newblock {\em Trends in cognitive sciences 22}, 9 (2018), 794--809.

\bibitem{parkhi2015deep}
{\sc Parkhi, O.~M., Vedaldi, A., and Zisserman, A.}
\newblock Deep face recognition.
\newblock In {\em British Machine Vision Conference\/} (2015).

\bibitem{phillips2014comparison}
{\sc Phillips, P.~J., and O'toole, A.~J.}
\newblock Comparison of human and computer performance across face recognition
  experiments.
\newblock {\em Image and Vision Computing 32}, 1 (2014), 74--85.

\bibitem{simonyan2014very}
{\sc Simonyan, K., and Zisserman, A.}
\newblock Very deep convolutional networks for large-scale image recognition.
\newblock {\em arXiv preprint arXiv:1409.1556\/} (2014).

\bibitem{srivastava2017training}
{\sc Srivastava, M., and Grill-Spector, K.}
\newblock Training a deep convolutional neural network with multiple face sizes
  and positions, but not resolutions, is necessary for generating invariant
  face recognition across these transformations.
\newblock {\em Journal of Vision 17}, 10 (2017), 247--247.

\bibitem{torralba2011unbiased}
{\sc Torralba, A., and Efros, A.~A.}
\newblock Unbiased look at dataset bias.
\newblock In {\em CVPR 2011\/} (2011), IEEE, pp.~1521--1528.

\bibitem{yamins2016using}
{\sc Yamins, D.~L., and DiCarlo, J.~J.}
\newblock Using goal-driven deep learning models to understand sensory cortex.
\newblock {\em Nature neuroscience 19}, 3 (2016), 356.

\bibitem{zeiler2014visualizing}
{\sc Zeiler, M.~D., and Fergus, R.}
\newblock Visualizing and understanding convolutional networks.
\newblock In {\em European conference on computer vision\/} (2014), Springer,
  pp.~818--833.

\end{thebibliography}

\end{document}